\documentclass[letterpaper]{article} % DO NOT CHANGE THIS
\usepackage{aaai25}  % DO NOT CHANGE THIS
\usepackage{times}  % DO NOT CHANGE THIS
\usepackage{helvet}  % DO NOT CHANGE THIS
\usepackage{courier}  % DO NOT CHANGE THIS
\usepackage[hyphens]{url}  % DO NOT CHANGE THIS
\usepackage{graphicx} % DO NOT CHANGE THIS
\usepackage{natbib}  % DO NOT CHANGE THIS AND DO NOT ADD ANY OPTIONS TO IT
\usepackage{caption} % DO NOT CHANGE THIS AND DO NOT ADD ANY OPTIONS TO IT
\usepackage{algorithm}
\usepackage{algorithmic}

\usepackage{newfloat}
\usepackage{listings}
\DeclareCaptionStyle{ruled}{labelfont=normalfont,labelsep=colon,strut=off} % DO NOT CHANGE THIS
\floatstyle{ruled}
\newfloat{listing}{tb}{lst}{}
\floatname{listing}{Listing}
\nocopyright % -- Your paper will not be published if you use this command
\usepackage{booktabs}
\usepackage{multirow}

\usepackage{color}

\usepackage{subcaption}

\title{FTII-Bench: A Comprehensive Multimodal Benchmark\\for Flow Text with Image Insertion}
\author{
    Jiacheng Ruan\textsuperscript{\rm 1}\equalcontrib,
    Yebin Yang\textsuperscript{\rm 1}\equalcontrib, 
    Zehao Lin\textsuperscript{\rm 2},
    Yuchen Feng\textsuperscript{\rm 2},
    Feiyu Xiong\textsuperscript{\rm 2},
    Zeyun Tang\textsuperscript{\rm 2},
    Zhiyu Li\textsuperscript{\rm 2\thanks{Corresponding author.}}
}
\affiliations{
    \textsuperscript{\rm 1} Shanghai Jiao Tong University 
    \\
    \textsuperscript{\rm 2} Institute for Advanced Algorithms Research, Shanghai. 
    \\
    jackchenruan@sjtu.edu.cn, lizy@iaar.ac.cn

}

\usepackage{bibentry}
\usepackage{amsmath}

\begin{document}

\maketitle

\begin{abstract}

% 我感觉上可以先论述这种新的任务，然后任务的重要性，然后提出了一个 benchmark 来测试，具体测试方式，最后当前主流模型/框架的测试结论，以及测试未来可以帮助模型发现哪些问题。

% 得益于大型语言模型（LLMs）和视觉基础模型的革命性进步，大型视觉语言模型（LVLMs）也取得了显著的发展。然而，目前的benchmark大多集中在物体识别、定位、图像理解和图文推理等任务上，这些任务往往仅考察LVLMs在某一单一方面的能力（如识别、检测、理解），未能充分展示它们在复杂应用场景中的潜力。为全面评估现有LVLMs的性能，本文提出了一项更具挑战性的任务，称为流式文本插图任务（Flow Text with Image Insertion Task，FTII task）。该任务要求LVLMs同时具备良好的图像理解、指令理解和长文本理解能力。具体而言，在给定若干文本段落和备选图像的情况下，随着文本段落的累积出现，LVLMs需要从备选图像中选出最适合的图像来插入相应段落之后。构建这种任务的benchmark极具挑战，尤其是在确定流式文本插图顺序时。为解决这一难题，我们转向专业新闻报道，这些报道天然包含金标准的图文顺序。基于此，我们引入了Flow Text with Image Insertion Benchmark（FTII-Bench），其中包含318条高质量中文图文新闻数据和307条高质量英文图文新闻数据，覆盖10个不同的新闻领域。我们利用这625条高质量数据，构建了两种不同类型及多级难度划分的问题。进一步地，我们基于CLIP模型和现有LVLM分别搭建了两种不同的评估pipeline。我们评估了9个开源和2个闭源的LVLMs及2个基于CLIP的模型。结果显示，即使是最先进的模型（如GPT-4o），在处理FTII任务时，仍面临巨大挑战。代码和数据样本可在附加材料中获取。
Benefiting from the revolutionary advances in large language models (LLMs) and foundational vision models, large vision-language models (LVLMs) have also made significant progress. However, current benchmarks focus on tasks that evaluating only a single aspect of LVLM capabilities (e.g., recognition, detection, understanding). These tasks fail to fully demonstrate LVLMs' potential in complex application scenarios. To comprehensively assess the performance of existing LVLMs, we propose a more challenging task called the Flow Text with Image Insertion task (FTII). This task requires LVLMs to simultaneously possess outstanding abilities in image comprehension, instruction understanding, and long-text interpretation. Specifically, given several text paragraphs and a set of candidate images, as the text paragraphs accumulate, the LVLMs are required to select the most suitable image from the candidates to insert after the corresponding paragraph. Constructing a benchmark for such a task is highly challenging, particularly in determining the sequence of flowing text and images. To address this challenge, we turn to professional news reports, which naturally contain a gold standard for image-text sequences. Based on this, we introduce the Flow Text with Image Insertion Benchmark (FTII-Bench), which includes 318 high-quality Chinese image-text news articles and 307 high-quality English image-text news articles, covering 10 different news domains. Using these 625 high-quality articles, we construct problems of two different types with multiple levels of difficulty. Furthermore, we establish two different evaluation pipelines based on the CLIP model and existing LVLMs. We evaluate 9 open-source and 2 closed-source LVLMs as well as 2 CLIP-based models. Results indicate that even the most advanced models (e.g., GPT-4o) face significant challenges when tackling the FTII task. The code and data will be available at https://github.com/IAAR-Shanghai/FTIIBench.

\end{abstract}
\section{Introduction}

% 注解：简介现有的LVLM
% 近年来，随着大型语言模型（LLM）的革命性进步以及视觉基础模型的提升，大型视觉语言模型（LVLM）取得了显著的发展。这些模型在图像理解、图文推理和视觉识别等任务中表现出卓越的性能，广泛应用于各个领域。例如，在医疗领域，LVLM被用于自动化病理图像分析和辅助诊断；在自动驾驶中，它们用于环境感知和场景理解。
In recent years, significant advancements in large visual-language models (LVLMs)~\cite{alayrac2022flamingo,liu2024visual,minigpt4,Dai2023InstructBLIPTG} have been achieved, driven by the revolutionary progress of large language models (LLMs) ~\cite{openai2023gpt,anil2023palm,touvron2023llama,llamamoe} and improvements in foundational visual models~\cite{vit,gan2022vision}. These models have shown outstanding capabilities in tasks such as image understanding, vision-language reasoning, and visual recognition, leading to widespread applications across various industries~\cite{singhal2023towards,wang2023drivemlm,mmcamobj}. For instance, in the medical field, LVLMs are employed for automated pathological image analysis and diagnostic assistance~\cite{singhal2023towards,li2024llava}, while in autonomous driving, they are utilized for environmental perception and scene understanding~\cite{xu2024drivegpt4,wang2023drivemlm}.

\begin{figure}
\centering
\includegraphics[width=0.92\linewidth]{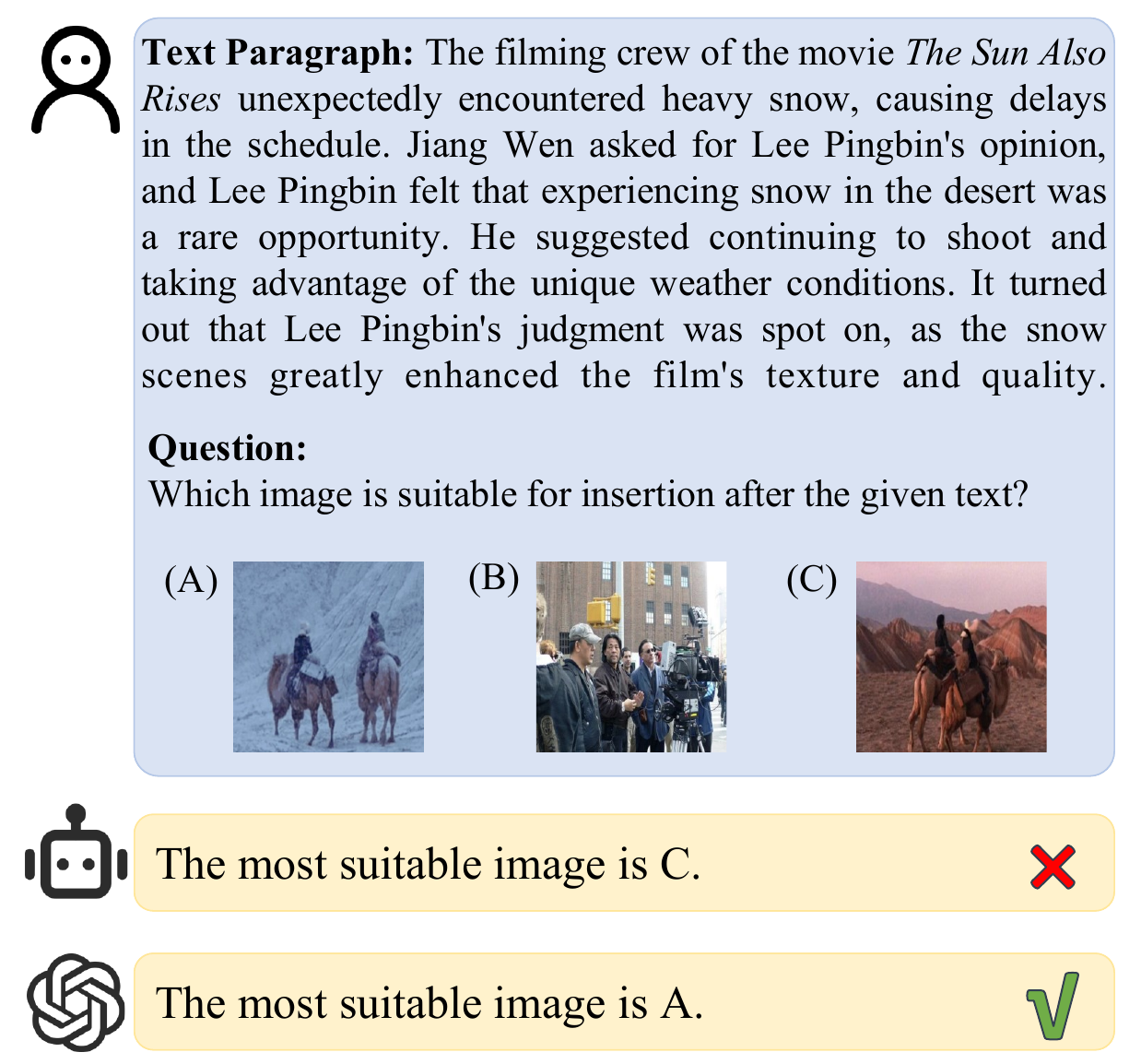}
\caption{An example of a single-choice question in our FTII-Bench.}
\label{fig:sc}
\end{figure}

% 随着越来越多的LVLMs被提出，如何对其进行全面评估成为当前的研究热点。然而，现有的基准测试主要集中在物体分类、定位及图文推理等任务，通常仅评估LVLMs在某一方面的能力（如识别、检测、理解），未能同时考察其在多个方面的综合能力。在实际应用中，LVLMs需处理更加动态的多模态信息。因此，为全面评估现有LVLM的性能，我们提出了一项更具挑战性的任务，即流式文本插图任务（Flow Text with Image Insertion Task，FTII task）。该任务包含多个文本段落和一个备选图像集合。文本段落以累加形式给出，LVLMs需从备选图像集合中选出最适合的图片插入到相应段落后。这对现有LVLMs同时提出了三个挑战：首先是图像理解能力，模型需充分理解图像集合中所有图片的含义；其次是长文本理解能力，由于文本是累加给出的，输入文本长度会不断增加；最后是指令遵循能力，要求LVLMs能按指令格式输出对所选图片的置信度。
With the increasing development of LVLMs, comprehensive evaluation of these models has become a research focus. However, existing benchmarks primarily focus on tasks such as object recognition, localization, and visual-text reasoning, usually evaluating only a single aspect of LVLMs' capabilities, such as recognition~\cite{liu2023hidden}, detection~\cite{fu2024blink}, or understanding~\cite{yang2021tap}, without simultaneously assessing their comprehensive abilities across multiple aspects. In real-world applications, LVLMs must handle more dynamic multimodal information. To thoroughly assess the performance of existing LVLMs, we propose a more challenging task, called the Flow Text with Image Insertion Task (FTII). This task comprises multiple text paragraphs and a set of candidate images. The text is provided cumulatively, requiring LVLMs to select the most suitable image from the candidates to insert after the corresponding paragraph. This presents three simultaneous challenges for current LVLMs: firstly, the ability to comprehend images, requiring the model to fully understand the meaning of all images in the set; secondly, the ability to understand long texts, as the cumulative nature of the text continuously increases the input length; and finally, the ability to follow instructions, demanding that LVLMs output the confidence level for the selected image in the specified format.

% 尽管构建这样一个benchmark充满挑战，尤其是在确定流式文本与插图顺序的答案时，但我们发现新闻领域的数据可以提供一个较为理想的金标准。在一些流媒体平台上，例如推特，图文内容的顺序往往缺乏专业性，因此难以用作可靠的基准。然而，在新闻报道中，尤其是官方新闻中，权威性和高质量的报道通常自然包含了准确的图文顺序。为此，我们手动从新华网和BBC新闻中收集了318条高质量的中文图文文章以及307条高质量的英文图文文章，涵盖政治、经济、科技等10个不同的新闻领域。基于这625条数据，我们设计了FTII-Bench，其中包含单选题和流式插入题两种类型。单选题分为四个难度等级，而流式插入题则分为三个难度等级。这种设计允许我们在不同层次上更全面地评估LVLMs的能力。
Constructing such a benchmark presents significant challenges, particularly in determining the sequence of streaming text and illustrations. However, we found that data from the news domain offers an ideal gold standard. On some multimedia platforms, such as Twitter, the order of text and images often lacks professionalism, making them unsuitable as a reliable benchmark. In contrast, news reports, especially official ones, inherently contain authoritative and high-quality sequencing of text and images. To this end, we manually collected 318 high-quality Chinese and 307 high-quality English image-text reports from Xinhua News and BBC News, covering ten different news domains such as politics, economics, and technology. Based on these 625 reports, we developed FTII-Bench, which includes both single-choice and flow-insertion question types. The single-choice questions are divided into four difficulty levels, while the flow-insertion questions have three difficulty levels. This design allows for a more comprehensive evaluation of LVLMs at different levels.

% 为了确保验证结果的全面性，我们不仅建立了LVLMs的评估流程，还引入了基于CLIP范式的评估流程。基于这些流程，我们在FTII-Bench上对9种开源LVLMs、2种闭源LVLMs以及2种基于CLIP的模型进行了充分验证。结果显示，即便是最先进的模型，在处理FTII任务时依然面临重大挑战。例如，最先进的GPT-4o在最困难的三选一单选题中仅获得了61.0\%的准确率。这表明，当前的LVLMs在应对复杂的多模态任务时仍有很大提升空间，需要进一步发掘其在多能力考验下的潜力。
To ensure the comprehensiveness of our validation results, we not only established an evaluation pipeline for LVLMs but also introduced an evaluation pipeline based on the CLIP~\cite{openaiclip} paradigm. Using these pipelines, we conducted extensive validation of 9 open-source LVLMs~\cite{llava,Dai2023InstructBLIPTG,yao2024minicpm,qwenvl,internlmxcomposer2_5,peng2023kosmos,idefics2,fuyu-8b}, 2 closed-source LVLMs, and 2 CLIP-based models~\cite{chen2024bge,BGE_en} on FTII-Bench. The results indicate that even the most advanced models face significant challenges when handling FTII tasks. For instance, GPT-4o achieved only 61.0\% accuracy on the most difficult single-choice questions (one out of three options). This suggests that current LVLMs still have substantial room for improvement in tackling complex multimodal tasks and that there is a need to further unleash their potential when tested on multiple capabilities simultaneously.

% 总的来说，本文的主要贡献如下：1）第一次地，我们提出了一个具有挑战性的新任务，FTII task，需要LVLMs同时具有长文本理解、图像理解以及复杂指令遵循的能力。2）我们手动收集了625条高质量新闻图文报道，并构建了包含10,231道问题的FTII-Bench，为后续的研究者提供了丰富的评估资源。3）我们评估了9种开源LVLMs，2种先进的闭源LVLM和以及2种基于CLIP的模型，揭示了这些模型在FTII任务中的性能瓶颈，并为未来的模型改进提供了方向。
% 总的来说，本文的主要贡献如下：1）我们首次提出了一项具有挑战性的任务，即FTII任务，该任务要求LVLMs具备长文本理解、图像理解和复杂指令遵循的能力。2）我们手动收集了625条高质量的新闻图文报道，并构建了包含10,231个问题的FTII-Bench，为后续研究提供了丰富的评估资源。3）我们评估了9种开源LVLM、2种先进的闭源LVLM和2种基于CLIP的模型，揭示了这些模型在FTII任务中的性能瓶颈，并为未来的模型改进指明了方向。

In summary, the main contributions of this paper are as follows: \textbf{1)} For the first time, We propose a novel and challenging task, dubbed the Flow Text with Image Insertion, which requires LVLMs to have the ability to understand long texts, comprehend images, and follow complex instructions. \textbf{2)} We manually collected 625 high-quality news image-text reports and constructed the FTII-Bench, which contains 10,231 questions, providing a rich evaluation resource for future researchers. \textbf{3)} We evaluated 9 open-source LVLMs, 2 advanced closed-source LVLMs, and 2 CLIP-based models, revealing the performance bottlenecks of these models in the FTII task and providing directions for future model improvements.
\section{Related Works}

\subsection{Large Vision and Language Models}
% 近年来，大型视觉语言模型（LVLMs）在多模态任务中取得了显著进展，成功地将视觉信息和文本信息进行了有效整合。Llava~\cite{llava} 通过利用大规模的图文数据集进行训练，在图像描述和视觉问答任务中表现出色。Qwen-VL~\cite{qwenvl} 通过一组可训练的查询提取并压缩视觉特征，并引入单层交叉注意力模块来实现视觉和文本模态的对齐，支持交织的图文输入。CogVLM~\cite{cogvlm} 通过在输入嵌入空间中将图像和文本拼接，并在语言模型中加入可训练的视觉层来实现两种模态的对齐。MiniCPM-V~\cite{yao2024minicpm} 通过有效的视觉token压缩和模态对齐提高了多模态模型的效率，在性能和计算成本之间实现了平衡，并且已经支持在多种端侧设备上部署。
In recent years, large vision language models (LVLMs) have achieved remarkable progress in multimodal tasks by effectively integrating visual and textual information. Llava~\cite{llava}, leveraging large-scale image-text datasets for training, excels in image captioning and visual question answering tasks. Qwen-VL~\cite{qwenvl} extracts and compresses visual features through a set of trainable queries and introduces a single-layer cross-attention module to achieve visual-text modality alignment, supporting interleaved image-text input. CogVLM~\cite{cogvlm} integrates images and text by concatenating them in the input embedding space and incorporates trainable visual layers within the language model to achieve the alignment between the two modalities. MiniCPM-V~\cite{yao2024minicpm} enhances the efficiency of multimodal models through effective visual token compression and modality alignment, achieving a balance between performance and computational cost. It is already supported for deployment on various edge devices.

% 参考intro的第二段，说一下虽然LVLMs的模型发展快速，但是大多数的benchmark，往往仅专注于单个方面的能力测试，（举上一两个bench的例子），这是不充分的。最后提一句我们的bench的优势。

%虽然LVLMs发展迅速，但如何全面地、同时地评估模型在多个方面的能力，仍有待充分的探究。目前有很多多模态benchmark涉及图文匹配和跨模态推理任务，但它们要么受限于有限的上下文长度\cite{mathew2021docvqa}，较少评估模型在理解长文本和多图像上下文中的能力。要么仅专注于单个视觉能力的评估，such as recognition~\cite{liu2023hidden}, OCR~\cite{masry2022chartqa}, detection~\cite{fu2024blink}, or understanding~\cite{yang2021tap}。这不足以评估模型在复杂任务以及实际应用中的表现。基于这种洞察，我们提出了FTII-Bench，能够同时考察模型的多个维度的能力，包括多图视觉理解、长上下文理解、复杂指令遵循等。
% 虽然LVLMs发展迅速，但如何全面且多维度地评估这些模型的能力仍有待深入探讨。目前，许多多模态基准测试涉及图文匹配和跨模态推理任务，但它们要么受到有限上下文长度的制约\cite{mathew2021docvqa,liu2023mmbench}，较少评估模型在理解长文本和多图像上下文中的能力，要么仅聚焦于单一视觉能力的评估，如识别~\cite{liu2023hidden}、OCR~\cite{masry2022chartqa}、检测~\cite{fu2024blink}或理解~\cite{yang2021tap}。这些评估方式难以全面反映模型在复杂任务和实际应用中的表现。因此，我们提出了FTII-Bench，能够同时评估模型在多图视觉理解、长上下文理解和复杂指令遵循等多个维度的能力。
Although LVLMs have rapidly evolved, a comprehensive and multi-dimensional evaluation of these models' capabilities remains underexplored. Currently, many benchmarks involve image-text matching and cross-modal reasoning tasks, but they are either constrained by single image or limited context length\cite{mathew2021docvqa,liu2023mmbench}, thereby inadequately assessing the models' ability to understand long textual and multi-image contexts, or they focus solely on evaluating individual visual capabilities, such as recognition~\cite{liu2023hidden}, OCR~\cite{masry2022chartqa}, math~\cite{lu2023mathvista}, detection~\cite{fu2024blink}, or understanding~\cite{yang2021tap}. These evaluation approaches are insufficient to fully reflect the models' performance in complex tasks and real-world applications. Therefore, we propose FTII-Bench, which simultaneously evaluates models across multiple dimensions, including multi-image visual understanding, long-context comprehension, and complex instruction following.

\subsection{CLIP-based Models}
% 介绍现有的CLIP-based model，例如openai clip，bge等
CLIP-based models have demonstrated remarkable capabilities in visual and language understanding in recent years. CLIP~\cite{openaiclip} utilizes contrastive learning on a large-scale dataset of image-text pairs, significantly enhancing the accuracy of matching between images and text while supporting zero-shot learning. This ability makes CLIP excel in tasks such as image-text retrieval and image classification, establishing it as an important foundational model for image-text understanding tasks.

To further improve the performance of multimodal models, ALIGN~\cite{jia2021scaling} model employs a similar contrastive learning strategy and leverages a larger dataset to optimize image-text representation learning. Florence~\cite{yuan2021florence} incorporates improved visual and language alignment techniques, optimizing the model's performance in visual understanding and text description tasks.

Building on CLIP, many studies have further explored its applications and optimizations for specific tasks~\cite{coop,maple,gist,lamm,idat}. For example, ViFi-CLIP~\cite{hanoonavificlip} enhances performance in video action recognition tasks by fine-tuning the CLIP. Additionally, BGE~\cite{chen2024bge,BGE_en} introduces a novel embedding model M3-Embedding, which excels in multilinguality, multifunctionality, and multi-granularity. It's capable of dense retrieval, multi-vector retrieval, and sparse retrieval, providing a unified model foundation for real-world IR applications.
% The Scene Text CLIP model focuses on scene text detection, enhancing the matching capability between text instances and visual features~\cite{yu2023turning}.

% \subsection{Image-Text Matching Task}
% 在多模态研究领域，近年来的进展得益于一系列关键的基准测试，这些基准测试旨在评估模型在图文匹配任务中的表现。MS COCO 数据集作为基础数据集，提供了大量带有详细描述的图像，被广泛用于图文检索和图像描述任务~\cite{lin2014microsoft}。VQA 数据集则通过要求模型基于视觉内容回答问题，挑战其在视觉与语言模态间的深度理解和推理能力~\cite{vqa}~\cite{vqa2}。与此同时，CLIP 基准测试是在 OpenAI 的 CLIP 模型基础上开发的，评估模型的跨模态检索能力，展示了 CLIP 在零样本学习和通过对比学习实现精准图文对齐方面的优势~\cite{radford2021learning}。CLIP 的图文检索基准特别关注基于 CLIP 模型的图文检索任务，通过大规模训练展示了增强的跨模态匹配能力。同样地，谷歌的 Conceptual Captions 数据集提供了一个开放领域的数据集，挑战模型通过利用网络数据理解和生成多样且新颖背景下的文本~\cite{sharma2018conceptual}。

% 这些基准测试涉及图文匹配和多场景设置中的跨模态推理任务。然而，它们通常仅限于单张图像和受限上下文，较少测试模型在理解长文本和多图像上下文中的能力。为了解决这一问题，我们的基准选择了一个精心策划的新闻数据集，通过流文本插图任务测试模型的上下文语义理解和多模态图文理解能力。

\section{Datasets}

\begin{figure*}[!t]
\centering
\includegraphics[width=0.87\linewidth]{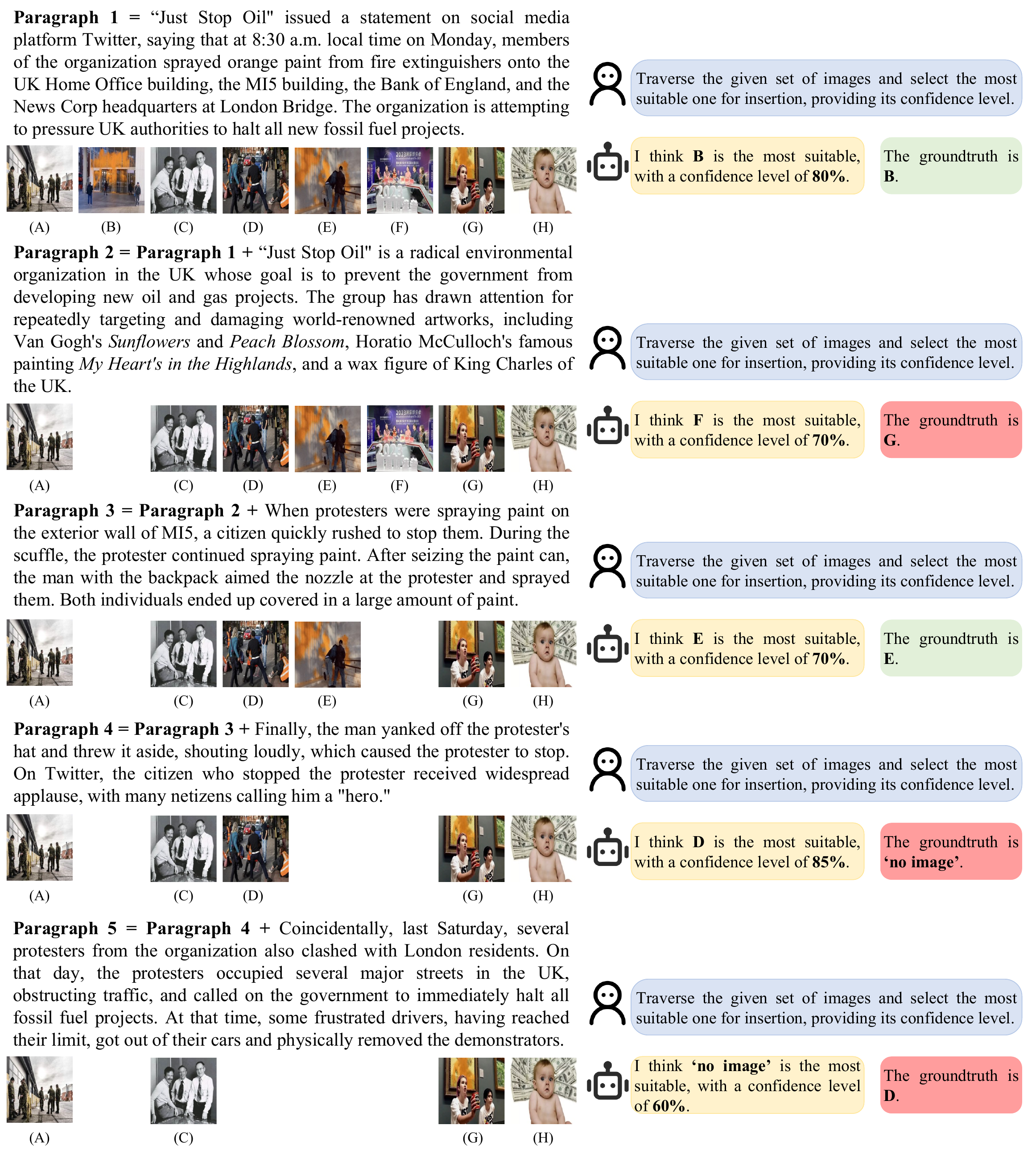}
\caption{An example of a flow-insertion question in FTII-Bench. In a flow-insertion question, news text paragraphs are provided cumulatively, and the corresponding ground truth images are presented alongside other distracting images as options.}
\label{fig:fi}
\end{figure*}

% 在本节中，我们详细介绍了FTII-Bench。首先，我们定义了Flow Text with Image Insertion任务。接着，我们描述了数据的收集过程以及相关统计信息。最后，我们说明了benchmark中的两种题型：单选题和流式插入题，并探讨了不同的难度设置。
In this section, we provide a detailed introduction to FTII-Bench. First, we define the Flow Text with Image Insertion Task. Next, we describe the data collection process and present relevant statistical information. Finally, we explain the two types of questions in the benchmark: single-choice questions and flow-insertion questions, and discuss the various difficulty settings.

\subsection{Task Definition}
% 给定一个图文交错的信息源$S$，其中包含$n$个文本段落和$m$张图片，即$S=\{s^t_1,...,s^t_n; s^i_1,...,s^i_m\}$。$S$遵循真实的图片插入顺序，记作$O_{gt}$。我们将groundtruth图片和文本分开，得到图片集合$I_s=\{s^i_1,...,s^i_m\}$和文本段落集合$T=\{s^t_1,...,s^t_n\}$。接着，我们引入包含$p$个干扰图片的集合$I_d=\{d^i_1,...,d^i_p\}$。通过将groundtruth图片集合与干扰图片集合合并，我们得到候选图片集合$I_c=\{s^i_1,...,s^i_m,d^i_1,...,d^i_p\}$。最后，我们将文本段落集合逐段累加作为LVLMs的文本输入，将候选图片集合逐一遍历作为图片输入，以便LVLMs提供置信度来确定图片的插入位置，最终得到预测的图片插入顺序$O_p$。任务的流程如Algorithm~\ref{alg:ftii}所示。
Given a source $\mathbf{S}$ that interleaves text and images, consisting of $n$ text paragraphs and $m$ images, i.e., $\mathbf{S} = \{s^t_1, \ldots, s^t_n; s^i_1, \ldots, s^i_m\}$. $\mathbf{S}$ follows a true image insertion order, denoted as $\mathbf{O_{gt}}$. By separating the ground truth images from the text, we can obtain the image set $\mathbf{I_s} = \{s^i_1, \ldots, s^i_m\}$ and the text paragraphs set $\mathbf{T} = \{s^t_1, \ldots, s^t_n\}$. At this point, a set of $p$ distractor images $I_d$ is introduced, i.e., $\mathbf{I_d} = \{d^i_1, \ldots, d^i_p\}$. Subsequently, by merging the ground truth image set and the distractor image set, we obtain the candidate image set $\mathbf{I_c} = \{s^i_1, \ldots, s^i_m, d^i_1, \ldots, d^i_p\}$. Finally, the text paragraphs set is incrementally accumulated as text input to the LVLMs, while the candidate image set is provided iteratively as image input. The LVLMs then provide confidence scores to determine the image insertion positions, ultimately resulting in the predicted image insertion order $\mathbf{O_p}$. The task workflow is illustrated in Algorithm~\ref{alg:ftii}.

\begin{algorithm}[!t]
\caption{ Flow Text with Image Insertion Task}
\label{alg:ftii}
\begin{algorithmic}[1]
\REQUIRE Text paragraphs $\mathbf{T} = \{s^t_1, \ldots, s^t_n\}$, ground truth images $\mathbf{I_s} = \{s^i_1, \ldots, s^i_m\}$, distracting images $\mathbf{I_d} = \{d^i_1, \ldots, d^i_p\}$, and a LVLM.
\ENSURE Predicted image-text order $\mathbf{O_p}$.
\STATE Initialize candidate image set $\mathbf{I_c} = \mathbf{I_s} \cup \mathbf{I_d} = \{s^i_1, \ldots, s^i_m, d^i_1, \ldots, d^i_p\}$.
\STATE Initialize empty order $\mathbf{O_p} \gets \{\}$.
\STATE Initialize accumulated text input $\mathbf{T}_{\text{accum}} \gets \{\}$.
\FOR{each text paragraph $s^t_k$ in $\mathbf{T}$}
    \STATE Add text paragraph $s^t_k$ to accumulated text $\mathbf{T}_{\text{accum}} \gets \mathbf{T}_{\text{accum}} \cup \{s^t_k\}$.
    \STATE Initialize maximum confidence score $\text{max\_score} \gets 0$.
    \STATE Initialize selected image $s^i_{\text{selected}} \gets \text{None}$.
    \FOR{each image $s^i_j$ in candidate image set $\mathbf{I_c}$}
        \STATE Provide image $s^i_j$ and accumulated text $\mathbf{T}_{\text{accum}}$ to the LVLM.
        \STATE Compute confidence score $c_j$ for image $s^i_j$.
        \IF{$c_j > \text{max\_score}$}
            \STATE Update $\text{max\_score} \gets c_j$.
            \STATE Update $s^i_{\text{selected}} \gets s^i_j$.
        \ENDIF
    \ENDFOR
    \STATE Append selected image $s^i_{\text{selected}}$ to order $\mathbf{O_p}$.
    \STATE Remove selected image $s^i_{\text{selected}}$ from candidate image set $\mathbf{I_c}$.
\ENDFOR
\RETURN $\mathbf{O_p}$
\end{algorithmic}
\end{algorithm}

% 在评估指标的计算中，我们考虑三种情况：仅计算插入图片的准确率（$\textup{Acc}_{i}$），仅计算未插入图片的准确率（$\textup{Acc}_{ni}$），以及同时考虑两者的准确率（$\textup{Acc}_{b}$）。为便于理解，以下是一个具体的例子：给定一个包含2张图片和5个文本段落的信息源，即$\mathbf{S}=\{s^t_1,s^t_2,s^i_1,s^t_3,s^t_4,s^t_5,s^i_2\}$，其图片插入的顺序为$\mathbf{O_{gt}}=\{\textup{None},\textup{None},s^i_1,\textup{None},\textup{None},s^i_2\}$。将图片和文本分离后，我们得到groundtruth图片集合$\mathbf{I_s}=\{s^i_1,s^i_2\}$及文本段落集合$\mathbf{T}=\{s^t_1,s^t_2,s^t_3,s^t_4,s^t_5\}$。此时，引入干扰图片集合$\mathbf{I_d}$，假设$\mathbf{I_d}$包含5张干扰图片，即$\mathbf{I_d}=\{d^i_1,d^i_2,d^i_3,d^i_4,d^i_5\}$。随后，将groundtruth图片集合与干扰图片集合合并，得到候选图片集合$\mathbf{I_c}=\{s^i_1,s^i_2,d^i_1,d^i_2,d^i_3,d^i_4,d^i_5\}$。最后，按照Algorithm~\ref{alg:ftii}的步骤，得到预测的图片插入顺序$\mathbf{O_p}$。假设$\mathbf{O_p}=\{\textup{None},d^i_2,s^i_1,\textup{None},\textup{None},d^i_5\}$。通过比较$\mathbf{O_{gt}}$与$\mathbf{O_p}$，我们可以计算出$\textup{Acc}_{i}=1/2=50\%, \textup{Acc}_{ni}=3/4=75\%, \textup{Acc}_{b}=4/7=57\%$。
When calculating evaluation metrics, we consider three scenarios: accuracy when only inserted images are considered ($\textup{Acc}_{i}$), accuracy when only non-inserted images are considered ($\textup{Acc}_{ni}$), and accuracy when both are considered ($\textup{Acc}_{b}$). For clarity, we present the following example: Given a source with 2 images and 5 text paragraphs, i.e., $\mathbf{S}=\{s^t_1,s^t_2,s^i_1,s^t_3,s^t_4,s^t_5,s^i_2\}$, the image insertion order follows $\mathbf{O_{gt}}=\{\textup{None},\textup{None},s^i_1,\textup{None},\textup{None},s^i_2\}$. By separating images and text, we obtain the groundtruth image set $\mathbf{I_s}=\{s^i_1,s^i_2\}$ and the text paragraph set $\mathbf{T}=\{s^t_1,s^t_2,s^t_3,s^t_4,s^t_5\}$. At this point, a distractor image set $\mathbf{I_d}$ is introduced, assuming $\mathbf{I_d}$ contains 5 distractor images, i.e., $\mathbf{I_d}=\{d^i_1,d^i_2,d^i_3,d^i_4,d^i_5\}$. The groundtruth image set is then merged with the distractor image set to form the candidate image set $\mathbf{I_c}=\{s^i_1,s^i_2,d^i_1,d^i_2,d^i_3,d^i_4,d^i_5\}$. Finally, following the steps in Algorithm~\ref{alg:ftii}, we obtain the predicted image insertion order $\mathbf{O_p}$. Assuming $\mathbf{O_p}=\{\textup{None},d^i_2,s^i_1,\textup{None},\textup{None},d^i_5\}$, by comparing $\mathbf{O_{gt}}$ and $\mathbf{O_p}$, we calculate $\textup{Acc}_{i}=1/2=50\%, \textup{Acc}_{ni}=3/4=75\%, \textup{Acc}_{b}=4/7=57\%$.

\subsection{Data Collection and Statistics}

% 在收集FTII任务相关数据的过程中，我们面临的主要挑战是如何准确确定图片的插入顺序。对于流媒体平台（如推特、微博）上的一般图文信息源而言，图片插入的顺序往往缺乏专业性，并且可能存在大量误导性信息。鉴于此，我们将研究重点转移到新闻领域。新闻网站（如新华网、BBC）通常包含大量图文交错的报道，且由于新闻报道的权威性，新闻中的图片插入顺序可以被视为可靠的标准。
The primary challenge in collecting data for the FTII task lies in accurately determining the sequence of image insertion. For general image-text sources on media platforms such as Twitter and Weibo, the sequence of image insertion often lacks professionalism and may contain a substantial amount of misleading information. Consequently, we have shifted our research focus to the news domain. News websites, such as Xinhua and BBC News, typically feature a large number of interwoven text and image reports, and due to the authoritative nature of news reporting, the image insertion sequence in news can be regarded as a reliable standard.

\begin{figure*}[!t]
    \centering
    \includegraphics[width=0.95\linewidth]{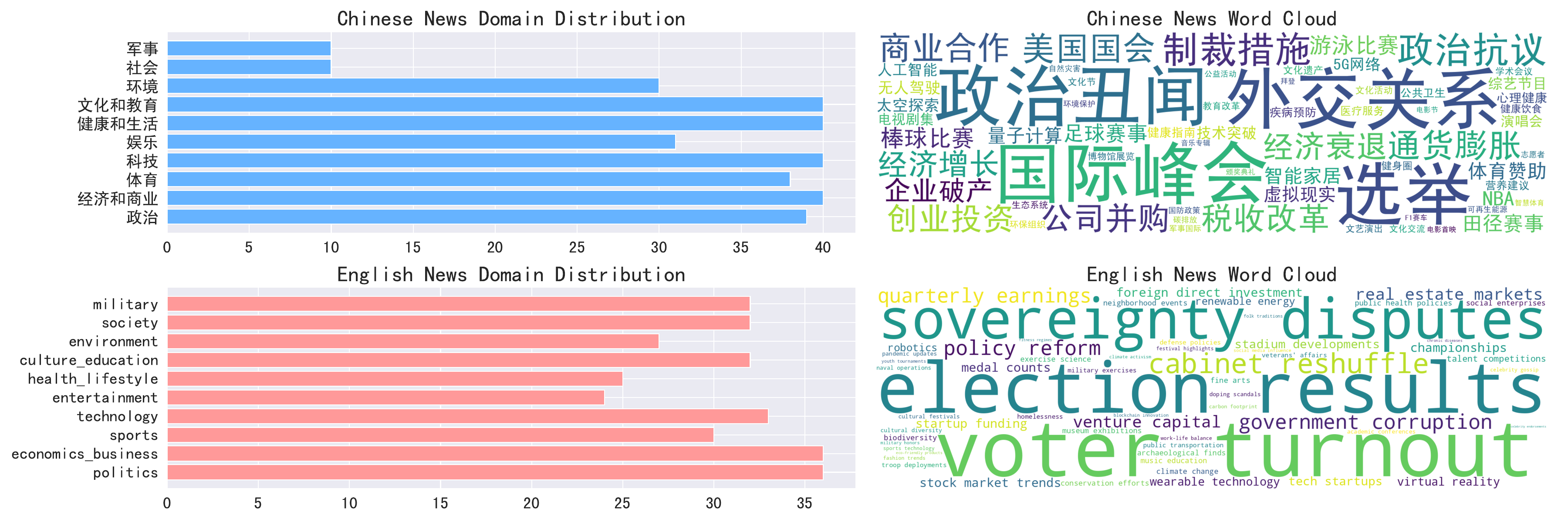}
    \caption{Data statistics regarding 318 high-quality Chinese news articles and 307 high-quality English news articles.}
    \label{fig:datastatistics}
\end{figure*}

% 为此，我们精心收集了来自新华网的318条高质量中文新闻和来自BBC的307条高质量英文新闻，涵盖了政治、经济与商业、体育、科技、娱乐、健康与生活、文化与教育、环境、社会及军事共10个领域。对于中文新闻，我们平均为每个领域指定了6.5个关键词，每个关键词下平均包含4.97条新闻。对于英文新闻，平均为每个领域指定了6个关键词，每个关键词下平均包含5.12条新闻。数据的统计信息如图~\ref{fig:datastatistics}所示。
To this end, we meticulously curated 318 high-quality Chinese news articles from Xinhua News and 307 high-quality English news articles from BBC, covering 10 domains: politics, economics and business, sports, technology, entertainment, health and lifestyle, culture and education, environment, society, and military. For the Chinese news, we assigned an average of 6.5 keywords per domain, with each keyword corresponding to an average of 4.97 articles. For the English news, we assigned an average of 6 keywords per domain, with each keyword corresponding to an average of 5.12 articles. The statistical information of the data is shown in Figure~\ref{fig:datastatistics}.

\subsection{FTII-Bench}

% 基于收集的图文新闻数据，我们构建了FTII-Bench。该基准测试包含两种题型：单选问题和流式插入问题。单选问题设置了四个难度等级，而流式插入问题则设计了三个不同的难度。
Based on the collected image-text news data, we construct FTII-Bench, which includes two types of tasks: single-choice questions and flow-insertion questions. The single-choice questions are designed with four levels of difficulty, while the flow-insertion questions feature three different levels of difficulty.

\subsubsection{Single-choice question} 
% 在我们收集的数据中，只有少数新闻的文本段落后附有图像。因此，对于单选题，我们仅使用这些段落后附带图像的文本作为问题的来源。随后，除了groundtruth图像之外，我们额外引入两个图像作为干扰，并通过控制干扰图像的来源，来调控single-choice question的难度等级。具体来说，我们将单选题分为四个难度等级，其中1级最简单，4级最难。4级难度：干扰图像来源于同一新闻条目下的其他图片。这些图像通常具有高度相似性，对模型的判断构成了较大的挑战。3级难度：干扰图像来自于同一领域和相同关键词下，但属于不同新闻条目的图片。这种设置增加了图像间的相似性，但不同新闻条目的背景知识有所变化。2级难度：干扰图像来源于同一领域下，但不同关键词和新闻条目的图片。这使得干扰图像在领域上相关，但主题和内容有所不同。1级难度：干扰图像来自不同领域、不同关键词和不同新闻条目的图片。这种设置下，干扰图像与问题本身的关联性最低，难度最小。通过这种干扰图像来源的精细化控制，我们可以更有效地评估模型在不同情境下的判别能力。如图 \ref{fig:sc}所示，为单选题的示例。

In our collected data, only a small portion of text paragraphs have the following image. Therefore, for the single-choice question, we use only those text paragraphs that are followed by images. Subsequently, in addition to the ground truth image, we introduce two additional images as distractors. By controlling the source of these distractor images, we adjust the difficulty levels of the single-choice question. Specifically, the questions are divided into four difficulty levels, with Level 1 being the easiest and Level 4 the hardest.
\begin{itemize}
    \item \textbf{Level 4 Difficulty:} Distractor images are sourced from the same news article. These images are usually highly similar, posing a significant challenge to the model's judgment.
    \item \textbf{Level 3 Difficulty:} Distractor images come from the same domain and keyword but are from different news articles. This increases the similarity among images, though the contextual background varies between articles.
    \item \textbf{Level 2 Difficulty:} Distractor images are from the same domain but have different keywords and news articles. This makes the distractor images related in domain but different in topic and content.
    \item \textbf{Level 1 Difficulty:} Distractor images are from different domains, keywords, and news articles. In this setup, the distractor images have the least relevance to the question, offering the lowest difficulty.
\end{itemize}
By finely controlling the sources of distractor images, we can more effectively assess the model's discriminative ability under different scenarios. As shown in Figure \ref{fig:sc}, this is an example of a single-choice question. 

\subsubsection{Flow-insertion question}

% 与单选题不同，流式插入题将多篇新闻文章中的图片组合成候选图片集合，如算法 \ref{alg:ftii} 中的 $I_c$ 所示。类似于单选题，流式插入题根据干扰图像的来源被划分为三个难度等级，难度从1级到3级逐渐增加。

% \begin{itemize}
% \item \textbf{3级难度：} 干扰图像来自相同领域并共享相同的关键词。由于图像之间的高度相似性，这种设置带来了最大的挑战。
% \item \textbf{2级难度：} 干扰图像来自相同领域但具有不同的关键词。这种变化通过改变上下文，同时保持领域相关性，引入了中等难度。
% \item \textbf{1级难度：} 干扰图像来自不同领域和关键词。由于干扰图像与问题的上下文几乎没有相关性，这提供了最简单的难度等级。
% \end{itemize}

% 这种分层的难度结构允许对模型在不同条件下辨别相关图像的能力进行细致的评估。如图 \ref{fig:fi}所示，为单选题的示例。

Unlike single-choice questions, Flow-insertion questions combine images from multiple news articles to form a set of candidate images, as represented by $I_c$ in Algorithm \ref{alg:ftii}. Similar to single-choice questions, Flow-insertion questions are divided into three difficulty levels based on the sources of the distractor images, with difficulty increasing from Level 1 to Level 3.
\begin{itemize}
    \item \textbf{Level 3 Difficulty:} Distractor images come from the same domain and share the same keywords. This setup presents the greatest challenge due to the high similarity among the images.
    \item \textbf{Level 2 Difficulty:} Distractor images are sourced from the same domain but have different keywords. This variation introduces moderate difficulty by altering the context while maintaining domain relevance.
    \item \textbf{Level 1 Difficulty:} Distractor images originate from different domains and keywords. This provides the easiest level, as the distractor images have minimal relevance to the context of the question.
\end{itemize}
This hierarchical structuring of difficulty levels allows for a nuanced evaluation of the model's ability to discern relevant images under varying conditions. As shown in Figure \ref{fig:fi}, this is an example of a Flow-insertion question. 

% 进一步地，我们分别对Single-choice questions以及Flow-insertion questions的不同难度进行了数据统计，如表格\ref{tab:benchstat}所示。
Furthermore, we conducted statistical analysis on the different difficulty levels of Single-choice questions and Flow-insertion questions separately, as shown in Tables \ref{tab:benchstat1} and \ref{tab:benchstat2}.

\begin{table}[!t]
\centering\small
\begin{tabular}{c|ccc}
\hline
\toprule
\textbf{Lan.} & \textbf{Level} & \textbf{\# Questions} & \textbf{Avg. Word Count} \\ 
\midrule
\multirow{4}{*}{CN} & 4 & 409 & 141.34 \\ 
                    & 3 & 907 & 153.01 \\ 
                    & 2 & 907 & 158.01 \\
                    & 1 & 843 & 153.51 \\ 
\midrule
\multirow{4}{*}{EN} & 4 & 711 & 35.60  \\ 
                    & 3 & 1079 & 25.73  \\ 
                    & 2 & 1084 & 35.14  \\ 
                    & 1 & 1032 & 35.83  \\ 
\bottomrule
\hline
\end{tabular}
\caption{Statistical analysis for single-choice questions.}
\label{tab:benchstat1}
\end{table}

\begin{table}[!t]
\centering
\footnotesize
\setlength{\tabcolsep}{2.5pt} % Adjust column separation
\begin{tabular}{c|cccc}
\hline
\toprule
\textbf{Lan.} & \textbf{Level} & \textbf{\# Questions} & \textbf{Avg. Word Count} &\textbf{Avg. \# Images}\\ 
\midrule
\multirow{3}{*}{CN} & 3 & 318 & 1093.35 & 13.95 \\ 
                    & 2 & 590 & 1130.89 &14.75 \\ 
                    & 1 & 749 & 1064.85 & 19.50 \\ 
\midrule
\multirow{3}{*}{EN} & 3 & 307 & 621.84 & 18.07 \\ 
                    & 2 & 630 & 629.56 &17.45 \\ 
                    & 1 & 665 & 609.23 & 24.13 \\ 
\bottomrule
\hline
\end{tabular}
\caption{Statistical analysis for flow-insertion questions.}
\label{tab:benchstat2}
\end{table}

\section{Experiments}

\begin{table*}[!t]
\centering
\setlength{\tabcolsep}{12.5pt} 
\begin{tabular}{lc|cc|cc|cc|cc}
\hline
\toprule
\textbf{Model} & \textbf{Size} & \multicolumn{2}{c|}{\textbf{Level 1}} & \multicolumn{2}{c|}{\textbf{Level 2}} & \multicolumn{2}{c|}{\textbf{Level 3}} & \multicolumn{2}{c}{\textbf{Level 4}} \\
&& \textbf{EN} & \textbf{CN} & \textbf{EN} & \textbf{CN} & \textbf{EN} & \textbf{CN} & \textbf{EN} & \textbf{CN} \\
\midrule
\hline
\multicolumn{10}{l}{\emph{Open-source LVLMs}}\\
\hline
 Llava-v1.5 &7B &34.6 &33.7 &33.5 &32.5 &33.5 &32.0 &31.6 &33.0 \\ 
 Llava-v1.5 &13B &33.9 &32.4 &34.2 &32.5 &32.3 &31.5 &32.8 &34.7 \\ 
 Idenfics2 &8B &31.9 &31.8 &33.1 &34.7 &32.7 &31.1 &35.4 &34.7 \\ 
  Fuyu &8B &32.7 &9.3 &33.2 &11.2 &33.1 &9.3 &33.2 &6.8 \\ 
 InstructBLIP &11B &35.3 &21.8 &33.5 &21.1 &31.6 &21.5 &33.1 &20.0 \\ 
 Kosmos2 &1.6B &35.4 &18.0 &32.9 &16.9 &33.8 &17.2 &33.2 &22.5 \\ 
 MiniCPM-V &8B &25.4 &33.1 &26.6 &34.8 &27.6 &32.2 &25.5 &30.1 \\ 
 Qwen-VL-Chat &7B &34.0 &33.3 &34.1 &29.7 &31.7 &31.5 &31.5 &33.5 \\ 
XComposer2.5 &7B &50.2 & 47.6 & 49.6 & 47.2 & 44.4 & 37.8 & 37.8 & 34.2 \\
\midrule
\hline
\multicolumn{10}{l}{\emph{Closed-source LVLMs}}\\
\hline
 GPT-4o & - &98.3 & 98.4 & 96.8 & 97.2 & 93.0 & 91.8 & 74.3 & 65.0 \\ 
 GPT-4o-mini & -  &96.3 & 97.7 & 94.5 & 94.2 & 86.4 & 85.0 & 63.6 & 57.0 \\
 % gemini-1.5-pro & 96.1 & 95.1 & 92.5 & &&&&53.7 \\
\midrule
\hline
\multicolumn{10}{l}{\emph{CLIP-based models}}\\
\hline
 BGE-M3 & - &90.6 &92.1 &85.0 &85.0 &78.1 &67.6 &55.0 &45.7 \\ 
 BGE-v1.5-en& -  &85.7 &- &82.6 &- &76.4 &- &50.1 &- \\ 
\bottomrule
\hline
\end{tabular}
\caption{The evaluation results of single choice questions in FTII-Bench: All questions are single-choice with three options, and accuracy (\%) is used as the evaluation metric.}
\label{tab:single_choice}
\end{table*}

% 在本节中，我们评估了现有开源与闭源的LVLMs，以及基于CLIP的模型在FTII-Bench上的表现。针对单选题目，我们将一段文本和三个图像选项同时输入到LVLMs中。由于存在多张发图片的同时输入，我们遵循\cite{mantis}中的设定，采用两种策略来处理这些图像：一种是将多张图像水平拼接成一张，另一种是将多张图像按序列输入模型。对于flow-insertion问题，我们首先给定一个文本段落，并采用逐一遍历图片集合的方式，将每张图片依次与文本段落组合，输入到LVLMs中。需要特别指出的是，每次输入仅包含一张图片。具体流程详见算法\ref{alg:ftii}。

In this section, we evaluate the performance of existing open-source and closed-source LVLMs, as well as CLIP-based models, on FTII-Bench. For single-choice questions, we input a text passage along with three image options into the LVLMs simultaneously. Due to the simultaneous input of multiple images, we follow the setup in \cite{mantis} and adopt two strategies to process these images: one strategy horizontally concatenates the images into a single image (`merge'), while the other inputs the images sequentially into the model (`sequence'). For the flow-insertion questions, we first provide a text paragraph and then sequentially traverse the image set, combining each image with the text paragraph as input to the LVLMs. It is important to emphasize that each input consists of only a single image. The detailed process is shown in Algorithm \ref{alg:ftii}.

\subsection{Baseline}
\subsubsection{Open-source LVLMs}

% 对于“合并”验证方式，我们使用了Fuyu-8B \cite{fuyu-8b}、InstructBLIP-FlanT5-XXL \cite{Dai2023InstructBLIPTG}、Kosmos2 \cite{peng2023kosmos}、Llava-v1.5-7B \cite{llava1_5}、Llava-v1.5-13B \cite{llava1_5}和Qwen-VL-Chat \cite{qwenvl}。对于“序列”验证方式，我们采用了Idenfics2 \cite{idefics2}、MiniCPM-Llama3-V-2.5 \cite{yao2024minicpm}和XComposer2.5 \cite{internlmxcomposer2_5}
For the `merge' validation method, we use Fuyu-8B \cite{fuyu-8b}, InstructBLIP-FlanT5-XXL \cite{Dai2023InstructBLIPTG}, Kosmos2 \cite{peng2023kosmos}, Llava-v1.5-7B \cite{llava1_5}, Llava-v1.5-13B \cite{llava1_5}, and Qwen-VL-Chat \cite{qwenvl}. For the `sequence' validation method, we adopt Idenfics2 \cite{idefics2}, MiniCPM-Llama3-V-2.5 \cite{yao2024minicpm}, and XComposer2.5 \cite{internlmxcomposer2_5}.

\subsubsection{Close-source LVLMs}

% 我们利用最先进的GPT-4o，并基于“序列”的方式来进行评估。
We employ the latest GPT-4o and GPT-4o-mini models, and conduct evaluations based on a `sequence' approach for single-choice questions.

\subsubsection{CLIP-based model}

% 相比于LVLMs，CLIP-based模型在指令遵循能力上有所不足。因此，在处理单选题时，我们首先给定一个文本段落，使用CLIP-based模型对该文本进行编码，生成对应的文本embedding。随后，我们对三个备选图像分别编码，生成相应的图像embedding，并计算它们与文本embedding之间的余弦相似度，选择得分最高的图像作为输出。在处理flow-insertion questions时，CLIP-based模型在获取文本embedding后，会遍历图像集合中的每一个图像，编码生成其embedding，并计算与文本embedding的余弦相似度，选择得分最高的图像作为输出。然而，需要注意的是，我们为flow-insertion questions引入了一个阈值（本文默认设置为0.5），仅当最高得分超过该阈值时，才将其作为输出，否则输出记为`None'。本文使用的CLIP-based模型包括BGE-M3 \cite{chen2024bge}和BGE-v1.5-en \cite{BGE_en}。
Compared to LVLMs, CLIP-based models lack the capability to follow instructions. Therefore, when handling single-choice questions, we first provide a text passage, which is encoded using a CLIP-based model to generate the corresponding text embedding. We then encode the three candidate images separately to generate their respective image embeddings and calculate the cosine similarity between these and the text embedding, selecting the image with the highest score as the output. For flow-insertion questions, after obtaining the text embedding, the CLIP-based model traverses each image in the set, encodes it to generate its embedding, and calculates the cosine similarity with the text embedding, selecting the one with the highest score as the output. However, it is important to note that for flow-insertion questions, we introduce a threshold (set to 0.5 by default in this paper), and only when the highest score exceeds this threshold is the image considered as the output; otherwise, the output is recorded as `None.' The CLIP-based models used in this paper include BGE-M3 \cite{chen2024bge} and BGE-v1.5-en \cite{BGE_en}.

\begin{table*}[!t]
\centering\footnotesize
\setlength{\tabcolsep}{1.5pt} 
\begin{tabular}{l|ccc|ccc|ccc||ccc|ccc|ccc}
\hline
\toprule
\multicolumn{1}{c|}{} & \multicolumn{9}{c||}{\textbf{EN}} & \multicolumn{9}{c}{\textbf{CN}}\\
\cmidrule{2-19}
 \textbf{Model} & \multicolumn{3}{c|}{\textbf{Level 1}} & \multicolumn{3}{c|}{\textbf{Level 2}} & \multicolumn{3}{c||}{\textbf{Level 3}} & \multicolumn{3}{c|}{\textbf{Level 1}} & \multicolumn{3}{c|}{\textbf{Level 2}} & \multicolumn{3}{c}{\textbf{Level 3}} \\
 & \textbf{$\textup{Acc}_{b}$} & \textbf{$\textup{Acc}_{i}$} & \textbf{$\textup{Acc}_{ni}$} & \textbf{$\textup{Acc}_{b}$} & \textbf{$\textup{Acc}_{i}$} & \textbf{$\textup{Acc}_{ni}$} &\textbf{$\textup{Acc}_{b}$} & \textbf{$\textup{Acc}_{i}$} & \textbf{$\textup{Acc}_{ni}$} & \textbf{$\textup{Acc}_{b}$} & \textbf{$\textup{Acc}_{i}$} & \textbf{$\textup{Acc}_{ni}$} & \textbf{$\textup{Acc}_{b}$} & \textbf{$\textup{Acc}_{i}$} & \textbf{$\textup{Acc}_{ni}$} & \textbf{$\textup{Acc}_{b}$} & \textbf{$\textup{Acc}_{i}$} & \textbf{$\textup{Acc}_{ni}$} \\
\midrule
\hline
\multicolumn{19}{l}{\emph{Open-source LVLMs}}\\
\hline
Llava-v1.5-7B &37.3 &2.7 &41.0 &44.1 &8.2 &49.9 &53.5 &2.9 &61.4 
&26.9 &11.1 &31.0 &37.4 &8.3 &45.2 &34.6 &5.7 &44.6\\ 
Llava-v1.5-13B &90.3 &0 &100.0 &86.1 &0 &100.0 &86.4 &0 &100.0 
&54.6 &2.2 &68.4 &44.4 &2.8 &55.6 &52.9 &2.9 &70.3\\ 
 Idenfics2  &36.0 &10.8 &38.7 &42.0 &9.8 &47.2 &52.3 &2.9 &60.1 
 &20.0 &2.2 &24.6 &28.1 &13.9 &31.9 &33.8 &22.9 &37.6\\ 
 Fuyu &65.0 &0 &72.0 &62.5 &1.6 &72.3 &64.7 &0 &74.9 
 &39.8 &4.4 &49.1 &46.2 &25.0 &51.9 &39.7 &11.4 &49.5\\ 
 InstructBLIP &51.4 &5.4 &56.4 &48.9 &3.3 &56.2 &60.9 &0 &70.4 
 &56.9 &4.4 &70.8 &60.2 &11.1 &73.3 &59.6 &5.7 &78.2\\ 
 Kosmos2 &82.5 &0 &91.3 &78.2 &1.6 &90.5 &79.5 &0 &91.9 
 &64.8 &4.4 &80.7 &67.2 &5.6 &83.7 &69.9 &17.1 &88.1\\ 
 MiniCPM-V &35.8 &0 &39.6 &41.4 &3.3 &47.5 &51.9 &0 &60.1 
 &22.2 &8.9 &25.7 &34.8 &29.2 &19.4 &31.9 &11.4 &38.6\\ 
 Qwen-VL-Chat &57.2 &8.1 &62.4 &50.1 &3.3 &58.6 &59.3 &2.9 &68.2 
 &35.2 &4.4 &43.4 &30.4 &16.7 &34.1 &35.3 &20.0 &40.6\\ 
XComposer2.5 &90.3&0&100.0&42.3&3.3&48.5&53.1&2.9&61.0 
&53.7&6.7&66.1&52.6&8.3&64.4&51.5&14.3&64.4\\
\midrule
\hline
\multicolumn{19}{l}{\emph{Closed-source LVLMs}}\\
\hline
 GPT-4o &81.7 &8.1 &89.6 &73.2 &8.2 &83.6 &70.9 &5.7 &81.2 
 &63.4 &8.9 &77.8 &56.7 &11.1 &68.9 &53.7 &25.7 &63.4\\ 
 GPT-4o-mini &83.1&10.8 &90.8 &75.9 &8.2 &86.8 &72.9 &0 &84.3 
 &53.2 &6.7 &65.5 &54.4 &16.7 &64.4 &55.1 &25.7 &65.3\\
\midrule
\hline
\multicolumn{19}{l}{\emph{CLIP-based models}}\\
\hline
 BGE-M3 &76.7 &6.6 &86.5 &75.1 &5.7 &85.1 &70.0 &2.9 &78.9 
 &61.4 &13.5&74.1 &58.7 &12.9 &70.7 &47.3 &12.6 &56.5\\
 BGE-v1.5-en &65.9 &6.0 &74.3 &64.6 &5.7 &73.1 &60.8 &1.9 &68.6 
 &- &- &- &- &- &- &- &- &- \\ 
\bottomrule
\hline
\end{tabular}
\caption{The evaluation results of flow-insertion questions in FTII-Bench.}
\label{tab:flow_insert}
\end{table*}

\subsection{Single-choice Question Results}

% 对于单选题，我们的设置是从三个选项中选择一个，即给定一段文本和三个备选图片，要求模型输出最适合插入的图像。在子数据集之间，根据图像来源的不同，其难度等级从1到4逐渐增加。
For single-choice questions, we set the task to select one out of three options: given a piece of text and three candidate images, the model is required to output the most suitable image to be inserted. Across different sub-datasets, the difficulty level gradually increases from 1 to 4 based on the source of the images.

% 表 \ref{tab:single_choice} 展示了基于 CLIP 的模型和 LVLMs 在 FTII-Bench 单选题上的评估结果。可以看到，在 CLIP-based 模型中，随着难度等级的提升，BGE 模型的准确率逐步下降，这表明通过控制干扰图像的来源来调节问题难度是一个有效的策略。在 closed-source LVLMs 中，GPT-4o 的表现最佳，在前三个难度等级中均取得了超过 90\% 的准确率。然而，在 Level 4 cn 难度下，GPT-4o 的准确率显著下降至 65.0\%，这表明我们的 benchmark 对现有的 LVLMs 构成了相当大的挑战，能够提出新的难题。
Table \ref{tab:single_choice} presents the evaluation results of CLIP-based models and LVLMs on FTII-Bench single-choice questions. It is observed that in CLIP-based models, the accuracy of the BGE model gradually decreases as the difficulty level increases, indicating that adjusting question difficulty by controlling the source of distractor images is an effective strategy. Among closed-source LVLMs, GPT-4o performs the best, achieving over 90\% accuracy in the first three difficulty levels. However, at Level 4 Chinese difficulty, GPT-4o's accuracy significantly drops to 65.0\%, demonstrating that our benchmark poses a considerable challenge to existing LVLMs, presenting new difficulties.

% 在开源LVLMs中，表现最为出色的模型是XComposer2.5-7B。在Level 1的中文和英文任务中，其表现分别达到了50.2\%和47.6\%，远超其他模型。即使是在相对较难的Level 3英文任务中，也取得了44.4\%的准确率。而其他开源LVLMs在四种难度的题目上，其准确率与随机选择相差无几。这种在多图同时输入场景下表现不佳的现象，也在\cite{fu2024blink,mantis}中得到了验证。一个可能的原因是这些模型在训练过程中，使用的多图输入模式和相关指令非常有限甚至完全没有，导致了它们在多图任务上的表现欠佳。与此不同，XComposer2.5-7B在训练过程中采用了Unified Dynamic Image Partition策略，该策略支持多种输入模式，包括文本、单图、多图及视频输入。这使得XComposer2.5-7B在应对复杂视觉任务时表现得尤为出色，特别是在多图场景下，其表现显著优于其他开源LVLMs。
Among open-source LVLMs, XComposer2.5-7B stands out as the best-performing model. It achieves accuracy rates of 50.2\% and 47.6\% on Level 1 tasks in Chinese and English, respectively, significantly outperforming other models. Even on the relatively challenging Level 3 English tasks, it manages an accuracy of 44.4\%. In contrast, other open-source LVLMs exhibit accuracy close to random guessing across four levels of difficulty. This poor performance in multi-image input scenarios is also observed in \cite{fu2024blink,mantis}. A possible reason for this underperformance is that these models were trained with limited or no use of multi-image input modes and related instructions, leading to suboptimal performance on multi-image tasks. Unlike others, XComposer2.5-7B employs the Unified Dynamic Image Partition strategy during training, which supports various input modes, including text, single-image, multi-image, and video inputs. This enables XComposer2.5-7B to excel in complex visual tasks, particularly outperforming other open-source LVLMs in multi-image scenarios.

\subsection{Flow insert Question Results}

% 在处理flow-insertion问题时，我们采用逐段累积的方式构建文本输入，并通过遍历候选图片集来确定最适合插入的图片，如图\ref{fig:fi}以及算法\ref{alg:ftii}所示。流式插入问题的评估结果如表\ref{tab:flow_insert_en}和\ref{tab:flow_insert_cn}所示。由英文和中文的评估结果可以看出，随着问题难度的增加，所有准确率指标在两种BGE模型中的表现均有所下降。例如，在英文任务上，BGE-v1.5-en的$\textup{Acc}_{b}$从Level 1难度的65.9\%下降至Level 2难度的64.6\%，进一步下降至Level 3难度的60.8\%；其他指标也呈现出相似的趋势。在中文任务上，BGE-M3的$\textup{Acc}_{b}$从Level 1难度的61.4\%下降至Level 2难度的58.7\%，再下降至Level 3难度的47.3\%。这进一步验证了通过控制干扰图像来源来调节问题难度的有效性。具体而言，随着问题难度的增加，groundtruth图像与干扰图像之间的相似度提高，导致基于CLIP的模型性能下降。
In addressing the flow-insertion questions, we adopt an incremental approach to construct text paragraphs, while determining the most suitable image for insertion by traversing the candidate image set, as shown in Figure \ref{fig:fi} and Algorithm \ref{alg:ftii}. The evaluation results of the flow-insertion questions are presented in Tables \ref{tab:flow_insert}. The results from both the English and Chinese tasks reveal a decline in accuracy across both BGE models as the task difficulty increases. For instance, in the English task, $\textup{Acc}_{b}$ of BGE-v1.5-en decreases from 65.9\% at Level 1 to 64.6\% at Level 2, and further to 60.8\% at Level 3; similar trends are observed for other metrics. In the Chinese task, $\textup{Acc}_{b}$ of BGE-M3 drops from 61.4\% at Level 1 to 58.7\% at Level 2, and further to 47.3\% at Level 3. This further validates the effectiveness of adjusting task difficulty by controlling the source of distractor images. Specifically, as the difficulty increases, the similarity between groundtruth and distractor images grows, leading to a performance decline in CLIP-based models.

% 对于LVLMs而言，flow-insertion问题不同于单选题，它不仅考验模型的图像理解能力，还涉及对长文本的理解以及复杂指令的遵循。现有的LVLMs在这一问题上的表现普遍不佳。例如，如表\ref{tab:flow_insert_en}所示，我们发现Llava-v1.5-13b模型在所有三个难度级别上的$\textup{Acc}_{i}$均为0\%，而$\textup{Acc}_{ni}$则为100\%。这表明Llava-v1.5-13b模型在理解是否应插入图像及其置信度方面存在严重问题，因此其每次选择的答案都是不需要插图。类似的表现也出现在其他模型上，如Fuyu-8b、Kosmos2和MiniCPM-V。这揭示了现有LVLMs在复杂指令理解和执行能力方面的局限性。此外，flow-insertion问题上的极差表现进一步说明了现有LVLMs在面对多重能力考验时的局限性。
For LVLMs, the flow-insertion questions, unlike single-choice questions, not only tests the model's image understanding capabilities but also challenges its comprehension of long texts and adherence to complex instructions. Current LVLMs generally perform poorly on this task. For instance, as shown in Table \ref{tab:flow_insert} EN, we observe that the Llava-v1.5-13B achieves an $\textup{Acc}_{i}$ of 0\% across all three difficulty levels, while its $\textup{Acc}_{ni}$ is 100\%. This indicates a significant deficiency in Llava-v1.5-13B's understanding of whether an image should be inserted and its confidence level, leading to the model consistently selecting answers that do not require an image. Similar issues are observed in other models, such as Fuyu-8b, Kosmos2, and MiniCPM-V. These findings reveal the limitations of current LVLMs in understanding and following complex instructions. Furthermore, the extremely poor performance on the flow-insertion questions further underscores the limitations of current LVLMs when faced with multiple simultaneous challenges.

\section{Conclusion}

% 在本文中，我们首次提出了Flow Text with Image Insertion任务，为现有的LVLMs带来了新的挑战。我们精心收集了625条双语数据集，其中包含318条高质量的中文图文新闻和307条高质量的英文图文新闻。在此基础上，我们引入了FTII-Bench，这一基准测试包括single-choice题和极具挑战性的flow-insertion题。单选题旨在考察模型的多图理解能力，而流式插入题则综合评估模型对图像、长文本及复杂指令的理解水平。实验结果显示，即便是最先进的闭源LVLMs，如GPT-4o，在流式插入题上仍然面临显著挑战。未来，我们计划引入更多数据进行模型微调，以进一步提升LVLMs在多重能力考验中的表现。
In this paper, we propose the Flow Text with Image Insertion task for the first time, introducing new challenges to existing LVLMs. We carefully compile a bilingual dataset of 625 entries, including 318 high-quality Chinese multimodal news items and 307 high-quality English multimodal news items. Based on these data, we introduce FTII-Bench, a benchmark comprising single-choice questions and highly challenging flow-insertion questions. The single-choice questions aim to assess the model's multi-image understanding capability, while the flow-insertion questions comprehensively evaluate the model's understanding of images, long texts, and complex instructions. Experimental results demonstrate that even the most advanced closed-source LVLMs, such as GPT-4o, still face significant challenges in flow-insertion tasks. In the future, we plan to incorporate more data to fine-tune the models, further enhancing LVLMs' performance when confronted with multiple competency evaluations.

\bibliography{aaai25}
\end{document}